\documentclass{eage}

\date{} % Certain latex templates automatically add the compilation date as a footnote in the generated pdf. This command can remove the date in some of the templates but does not work for all.

\usepackage{enumitem}

\usepackage{url}

\usepackage{hyperref} 
\usepackage{amsmath, graphicx}
\usepackage{multirow}
\usepackage{caption}
\usepackage{amsopn}
\newcommand{\suchthat}{\;\ifnum\currentgrouptype=16 \middle\fi|\;}
\DeclareMathOperator{\Fscore}{\textit{F-score}}
\providecommand{\keywords}[1]{\textbf{\textit{Keywords---}} #1}
\usepackage{setspace}
\begin{document}

%%%%%%%%% for cover page
\onecolumn % make sure you keep this coverpage as one column. In this template, we force the coverpage to be one column with this command and then switch to double column for the remaining of the paper with the \doublecolumn command. 

\begin{description}[labelindent=0.5cm,leftmargin=5cm,style=multiline]

\item[\textbf{Citation}]{M. Shafiq, T. Alshawi, Z. Long, and G. AlRegib, ``The role of visual saliency in the automation of seismic interpretation,'' Geophysical Prospecting, vol. 66, issue S1, pp. 132-143, Mar. 2018.}

\item[\textbf{DOI}]{\url{https://doi.org/10.1111/1365-2478.12570}}

\item[\textbf{Review}]{Date of publication: 15 September 2017}

\item[\textbf{Data and Codes}]{\url{https://ghassanalregibdotcom.files.wordpress.com/2016/10/amir_gp2017_code.zip}}% If you do not have data related to this paper, you can remove the data keyword.

\item[\textbf{Bib}] {@article\{shafiq2018role,\\
  title=\{The role of visual saliency in the automation of seismic interpretation\},\\
  author=\{Shafiq, M. and Alshawi, T. and Long, Z. and AlRegib, G.\},\\
  journal=\{Geophysical Prospecting\},\\
  volume=\{66\},\\
  number=\{S1\},\\
  pages=\{132--143\},\\
  year=\{2018\},\\
  month=\{Mar.\},\\
  publisher=\{Wiley Online Library\}\}
} 

% Preprint sharing policy can vary depending on the publisher. Before posting a paper to arXiv, please specifically check the transaction/conference you are targeting. In some transactions, papers are usually added to arxiv after acceptance. Publishers usually allow the authors to share accepted version of their papers but not the final formatted version that is provided by the publisher. In case of sharing preprints, publishers usually ask to add DOI and citation to the paper along with a copyright notice.

\item[\textbf{Copyright}]{\textcopyright 2017 European Association of Geoscientists \& Engineers (EAGE). Personal use of this material is permitted. Permission from EAGE must be obtained for all other uses, in any current or future media, including reprinting/republishing this material for advertising or promotional purposes, creating new collective works, for resale or redistribution to servers or lists, or reuse of any copyrighted component of this work in other works.}

\item[\textbf{Contact}]{\href{mailto:zhiling.long@gatech.edu}{zhiling.long@gatech.edu}  OR \href{mailto:alregib@gatech.edu}{alregib@gatech.edu}\\ \url{https://ghassanalregib.com/} \\ }
\end{description}

% Following command sequence was used to start the paper content from the following page and avoid numbering cover page.
\thispagestyle{empty}
\newpage
\clearpage
\setcounter{page}{1}

% Cover page was 1 column. \twocolumn changes the page format back to double column.
%\twocolumn
%%%%%%%%% for cover page

\title{The Role of Visual Saliency in the Automation of Seismic Interpretation}

\renewcommand{\thefootnote}{\fnsymbol{footnote}}

\author{Muhammad Amir Shafiq\footnotemark[1]$^{\dagger}$, Georgia Institute of Technology, and Tariq Alshawi, Georgia Institute of Technology, and Zhiling Long$^{\dagger}$, Georgia Institute of Technology, and Ghassan AlRegib$^{\dagger}$, Georgia Institute of Technology,.}

\footer{Geophysical Prospecting}
\lefthead{Shafiq et al.}
\righthead{Saliency in seismic interpretation}

\maketitle

\begin{abstract}
In this paper, we propose a workflow based on \emph{SalSi} for the detection and delineation of geological structures such as salt domes. \emph{SalSi} is a seismic attribute designed based on the modeling of human visual system that detects the salient features and captures the spatial correlation within seismic volumes for delineating seismic structures. Using \emph{SalSi}, we can not only highlight the neighboring regions of salt domes to assist a seismic interpreter but also delineate such structures using a region growing method and post-processing. The proposed delineation workflow detects the salt-dome boundary with very good precision and accuracy. Experimental results show the effectiveness of the proposed workflow on a real seismic dataset acquired from the North Sea, F3 block. For the subjective evaluation of the results of different salt-dome delineation algorithms, we have used a reference salt-dome boundary interpreted by a geophysicist. For the objective evaluation of results, we have used five different metrics based on pixels, shape, and curvedness to establish the effectiveness of the proposed workflow. The proposed workflow is not only fast but also yields better results as compared to other salt-dome delineation algorithms and shows a promising potential in seismic interpretation.
\end{abstract}

\newpage
\keywords{Saliency, \emph{SalSi}, Seismic attribute, Salt dome delineation, Seismic interpretation, Data processing, Signal processing.}

\newpage

\section{Introduction}

The evaporation of water from a basin causes the deposition of salt evaporites. Over a long periods of time, these evaporites, because of their low density, break through sediment layers often composed of limestone and shale to form diapir-shaped structures called salt domes. Salt domes may span over several kilometers in the Earth's subsurface and form stratigraphic traps for petroleum and gas reservoirs because of their impermeability. Therefore, accurate localization and delineation of the salt domes in a migrated seismic volume is one of the key steps in the exploration of oil and petroleum reservoirs. Experienced interpreters can manually label the boundaries of salt domes by observing and analyzing seismic reflections. However, with the dramatically increasing size of seismic data, manual labeling is becoming extremely time consuming and labor intensive. In recent decades, to improve interpretation efficiency, researchers have used intelligent computer-aided algorithms to assist the interpretation process. Interpreters beginning with an initial solution can interactively fix the erroneously detected boundary sections and fine tune the algorithm's parameters to accurately segment seismic volumes. Therefore, fully- and semi-automated algorithms for seismic interpretation under the supervision of interpreters have proved their worth both in industry and academia.

Over the last few decades, researchers have proposed several subsurface structure detection methods. In particular, there are several works on the detection of salt domes such as edge-based methods by \citealt*{jing2007detecting, aqrawi2011detecting}; and \citealt*{asjadmultiedge}, texture-based methods by \citealt*{berthelot2013texture, Zhen2015GoT}; and \citealt*{Shafiq2015GoT}, graph-theory-based method by \citealt*{shi2000normalized}, active-contours-based detection by \citealt*{winston_LSE, lse_schlum}; and \citealt*{Shafiq_AC}, machine learning-based methods by \citealt*{Pablo_texture, asjad2015hybrid, Pablo_ml, Pablo_novel}; and \citealt*{asjadcodebook}, and different image processing techniques by \citealt*{lomask2004image, lomask2007application, halpert2009seismic, Qi_Marfurt_SDMTC_2016, ramirez2016salt}; and \citealt*{Wu_salt_2016}. One of the rarely explored aspect for seismic interpretation is saliency.

Saliency detection attempts to predict areas in images and videos that are interesting to humans typically called salient regions by relying on low level features that attracts the human visual system (HVS)~(Borji and Itti 2013). As a great deal of research in computational cognitive science suggest, HVS has evolved to reduce the size of the sensory data information gathering stage, also known as the task-free visual search, by focusing on the perceptually salient segments of visual data that convey the most useful information about the scene~(Borji 2015). Features like color contrast, intensity contrast, flicker, and motion all have been identified as prominent features that help HVS to focus processing resources on important elements in the surrounding environment. It is commonly believed that HVS is attracted to localized outliers and novel elements in the environment, which is formulated as center-surround model by \citealt*{Gao08}. The center-surround model compares regions in the visual input to its local surrounding and predicts its saliency. Several features and detection algorithms for saliency have been proposed in the literature by \citealt*{Borji2013} and \citealt*{Borji2015}. More recently, a 3D FFT-based saliency detection algorithm for videos has been proposed by \citealt*{Long2015}. This algorithm uses a 3D FFT of a non-overlapping window in the spatial and temporal domains of a video sequence to compute the spectral energy of the window and compare it with its surrounding regions to construct the saliency map. The algorithm is efficient computationally, requires little tuning of model parameters as compared to other visual saliency algorithms, and provides reliable results as shown in~\citealt*{Long2015}.

In seismic interpretation, visual saliency is important to predict the human interpreters attention and highlight the areas of interest in seismic sections. \citealt*{Drissi_Sal} proposed an algorithm for horizon picking by detecting the salient texture features in seismic sections by computing entropy at each pixel using two entropy measures: the Shannon entropy and the generalized cumulative residual entropy. After saliency detection, \citealt{Drissi_Sal} used active contour for tracking horizons in seismic volume. On the other hand, \citealt*{Shafiq_sal} proposed a seismic attribute for salt dome detection based on visual saliency, \emph{SalSi}. \emph{SalSi} highlights the salient areas of a seismic image, i.e. the neighborhood of salt-dome boundaries, by comparing local spectral features based on 3D fast Fourier transform (FFT) according to \citealt*{Long2015}. However, to the best of our knowledge, saliency has not been proposed for salt dome delineation.

In this paper, we propose a workflow based on \emph{SalSi} for salt dome delineation, which is a continuation of our previous work (Shafiq \emph{et~al}. 2016a,b), where we proposed \emph{SalSi} for seismic interpretation. Using the proposed workflow, we can process seismic volumes in real-time and perform complex processing procedures more precisely. The rest of the paper is organized as follows. The proposed workflow for salt dome delineation is presented in section 2. Experimental results on a 3D field data are given in section 3 followed by conclusions in section 4.

\begin{figure}
  \centering
  \includegraphics[width=\columnwidth]{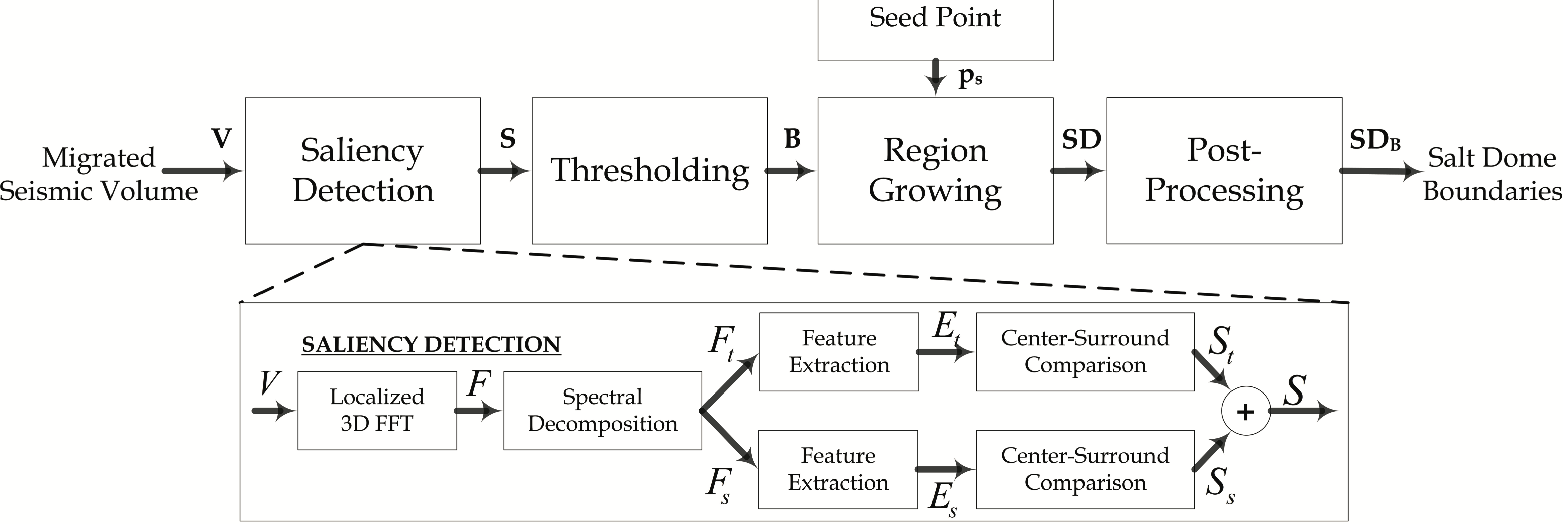} 
  \caption{The block diagram of the proposed workflow.}\label{BD_Del}
\end{figure}

\section*{Proposed Workflow for Salt Dome Delineation}
\label{sec:proposed}
The block diagram of the proposed workflow for salt dome delineation is shown in Figure~\ref{BD_Del}. The migrated 3D seismic data, $\mathbf{V}$, is of size $M \times N \times K$, where $M$ represents the number of samples of time or depth axis, $N$ represents the number of crosslines, and $K$ represents the number of inlines. There are four main steps of the proposed workflow as explained in detail below.

\subsection{Saliency Detection}
We, first, compute saliency using the 3D FFT-based algorithm as proposed in~\citealt*{Long2015}. The 3D FFT-based saliency algorithm obtains saliency maps with adjustable resolution, which allows better segmentation of salient objects. The 3D FFT-based algorithm is computationally inexpensive and requires little tuning of model parameters as compared to other visual saliency algorithms, which make it advantageous for seismic applications. The block diagram of 3D FFT-based saliency detection algorithm is also shown in Figure~\ref{BD_Del}. To obtain the saliency map $\mathbf{S}$, we calculate the 3D FFT spectrum $\mathbf{F}$ in a local area using equation (\ref{eqn:3dfft}), and decompose $\mathbf{F}$ into a temporal-change-related component $\mathbf{F_t}$ and a spatial-change-related component $\mathbf{F_s}$ as
\begin{equation} \label{eqn:3dfft}
\mathbf{F}[\mu,\nu,\omega]=\frac{1}{L^3}\sum\limits_{m=0}^{L-1}\sum\limits_{n=0}^{L-1}\sum\limits_{k=0}^{L-1}f[m,n,k]e^{-2\pi i\left(m \mu + n \nu + k \omega \right)/L},
\end{equation}
\begin{equation}
\mathbf{F_t}[\mu,\nu,\omega]=\mathbf{F}[\mu,\nu,\omega] \times \frac{\omega}{\sqrt[]{\mu^2+\nu^2+\omega^2}},
\end{equation}

\begin{figure}[t]
  \centering
 % \vspace{0.7cm}
  \includegraphics[width=.7\columnwidth]{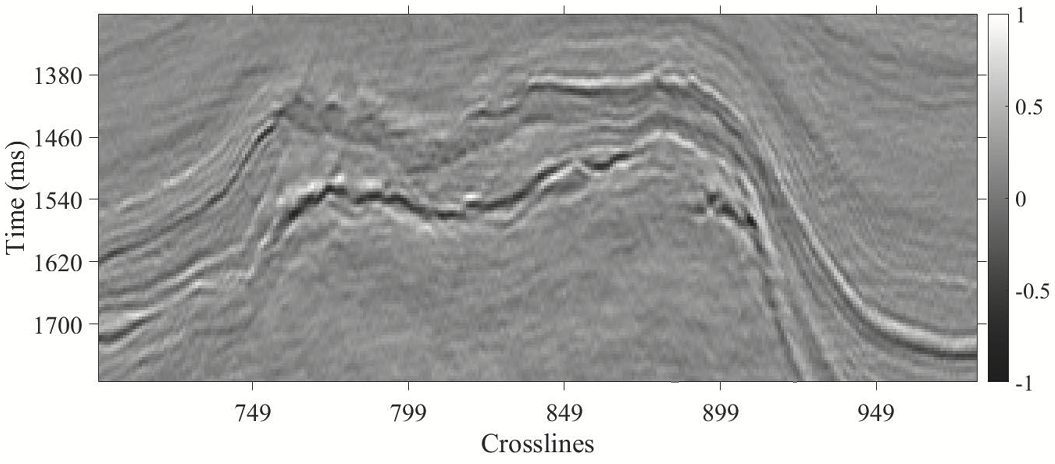}
  \caption{A typical seismic inline containing salt dome.}\label{SS160_orig}
  \vspace{0.4cm}
\end{figure}
\begin{figure}[t]
  \centering
  \includegraphics[width=.7\columnwidth]{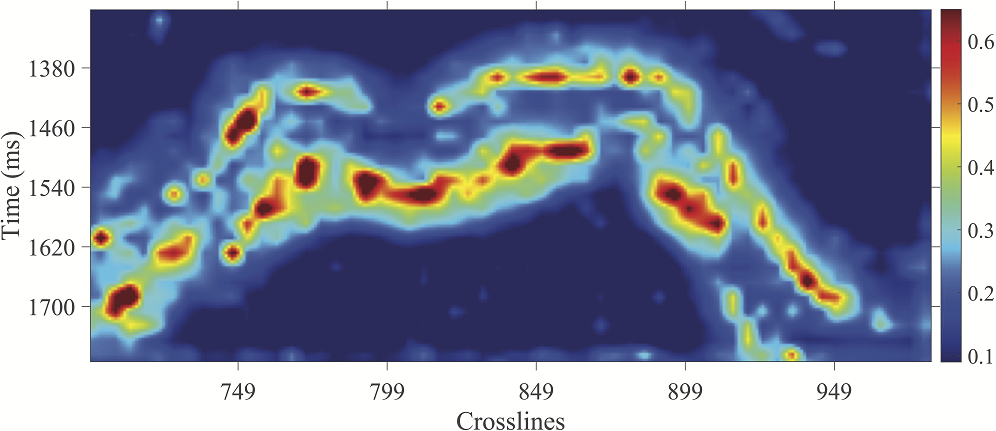} 
  \caption{The saliency map of a seismic inline.}\label{SS160_Salmap}
\end{figure}


\begin{equation}
\mathbf{F_s}[\mu,\nu,\omega]=\mathbf{F}[\mu,\nu,\omega] \times \frac{\sqrt[]{\mu^2+\nu^2}}{\sqrt[]{\mu^2+\nu^2+\omega^2}},
\end{equation}
\noindent where $[m,n,k]$ and $[\mu,\nu,\omega]$ represent the coordinates in the spatial and frequency domains, respectively, $L$ defines the size of local data cube, and $f[m,n,k]$ is the seismic image or section. Subsequently, the spectral energies $\mathbf{E_t}$ and $\mathbf{E_s}$ are calculated as features based on absolute mean of temporal- and spatial-change-related components as
\begin{align}\label{eqn:E_x}
% \mathbf{E_x}[m,n,k]= mean(\mathbf{F_x}), \quad x \in (t,s),
\mathbf{E_x}[m,n,k]= \frac{1}{L^3} \sum_{i,j,k}\mathbf{|F_x|}, \quad x \in (t,s),
\end{align}
where $\mathbf{F_x}$ represents the local spectral volume centered around a voxel $[m,n,k]$. Applying the center-surround model, two saliency maps $\mathbf{S_t}$ and $\mathbf{S_s}$ can be constructed using $\mathbf{E_t}$ and $\mathbf{E_s}$ as
\begin{align}\label{eqn:S_x}
\mathbf{S_x}[m,n,k]= \frac{1}{Q} \sum_{i_0,j_0,r_0} | \mathbf{E_x}[m,n,k] - \mathbf{E_x}[m+i_0,n+j_0,k+r_0] |,
\end{align}
\noindent where $i_0$, $j_0$, $r_0$ are chosen such that point $[m+i_0,n+j_0,k+r_0]$ is in the immediate neighborhood of point $[m,n,k]$, such as within a $3 \times 3 \times 3$ window centered at $[m,n,k]$. $\mathbf{S_x}$ represents $\mathbf{S_t}$ or $\mathbf{S_s}$, $\mathbf{E_x}$ represents $\mathbf{E_t}$ or $\mathbf{E_s}$, and $Q$ represents the total number of points included in the summation in equation~\ref{eqn:S_x}. The final saliency map $\mathbf{S}$ is obtained by averaging $\mathbf{S_t}$ and $\mathbf{S_s}$, and is of same size as of $\mathbf{V}$.
\begin{equation} \label{eqn:S}
\mathbf{S}[m,n,k]=0.5 \times \mathbf{S_t}[m,n,k] + 0.5 \times \mathbf{S_s}[m,n,k].
\end{equation}
A typical seismic inline section and its normalized saliency map $\mathbf{S}$ are shown in Figures~\ref{SS160_orig} and \ref{SS160_Salmap}, respectively. It can be observed from the Figure~\ref{SS160_Salmap} that it highlights the boundary of salt dome, which can be extracted using the following steps of the proposed workflow.

\subsection{Thresholding}
\begin{figure}[t]
  \centering
 % \vspace{0.7cm}
  \includegraphics[width=.7\columnwidth]{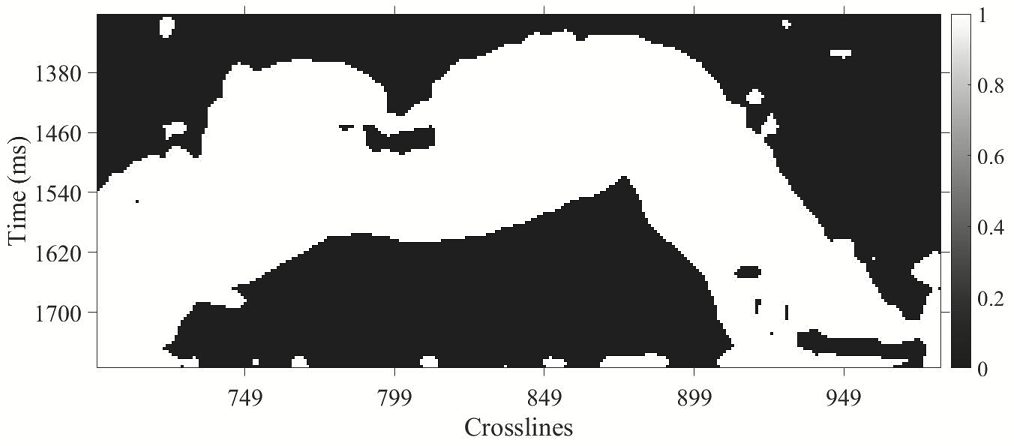}
  \caption{The binary map of a seismic inline.}\label{SS160_Binmap}
\end{figure}

For the application under consideration, the most salient part of a seismic image is the salt-dome boundary as seen in the saliency map $\mathbf{S}$. The second step of the proposed workflow is to threshold the saliency map to obtain a binary volume $\mathbf{B}$, which highlights the salt-dome boundaries.
\begin{equation}
\label{eqn:thres}
\mathbf{B}[m,n,k]=
\left\{
\begin{aligned}
&1\quad \mathbf{S}[m,n,k]\geq \mathbf{T}\\
&0\quad \mbox{Otherwise}
\end{aligned}
\right.
,
\end{equation}
\noindent where $\mathbf{T}$ represents the threshold. In contrast to the non-salt regions, salt-dome boundaries have higher $\mathbf{S}$ values. Therefore, we assume that the histogram of the volume $\mathbf{S}$ follows a bi-modal distribution. The threshold $\mathbf{T}$ can be determined by minimizing the intra-class variance by optimally dividing all points into two classes. Mathematically, it can be written as
\begin{equation}
\label{equ:otsu}
T = \arg\underset{T}{\min}\left\{\sigma_{1}^2(T)\sum_{i=0}^{T-1}p(i)+\sigma_{2}^2(T)\sum_{i=T}^{H}p(i)\right\},
\end{equation}
\noindent where $H$ is the number of the quantized gray-levels of $\mathbf{S}$, and $p(i)$, $i=0,\cdots,H-1$, represents the probability of points with gray value $i$. In addition, $\sigma_1^2$ and $\sigma_2^2$ define the individual class variances, which can be calculated as follows:
\begin{equation}
\label{equ:var}
\left\{
\begin{aligned}
&\sigma_1^2=\sum\limits_{i=0}^{T-1}\left[i-\sum\limits_{i=0}^{T-1}\frac{ip(i)}{p_1}\right]^2\frac{p(i)}{p_1},\quad p_1=\sum\limits_{i=0}^{T-1}p(i)\\
&\sigma_2^2=\sum\limits_{i=T}^{H-1}\left[i-\sum\limits_{i=T}^{H-1}\frac{ip(i)}{p_2}\right]^2\frac{p(i)}{p_2},\quad p_2=\sum\limits_{i=T}^{H-1}p(i)\\
\end{aligned}
\right.
.
\end{equation}
Therefore, we can adaptively identify threshold $\mathbf{T}$ by exhaustively searching between $0$ and $H-1$. Otsu's method attempts to find a value $\mathbf{T}$ by iteratively searching over the possible probability values such that the interclass variance are maximized. Otsu shows in \citealt*{Otsu_method} that the optimum threshold $\mathbf{T}$ can be obtained by maximizing the inter-class variance which is equivalent to minimizing the intra-class variance. Mathematically, it is represented as
\begin{equation}
% T = \arg \max_{T} \left\{ \left(\sum_{i=0}^{T-1}{p(i)}\right)     \left(\sum_{i=0}^{T-1}{p(i)} \right)       \left(\mu_1(T) - \mu_2(T)\right) \right\} ,
T = \arg \max_{T} \left\{ \left(\sum_{i=0}^{T-1}{p(i)} \right)       \left(\mu_1(T) - \mu_2(T)\right) \right\} ,
\end{equation}
where $\mu_1(T)$ and $\mu_2(T)$ are the mean values of the first and second classes, respectively, at a threshold $T$. In the quest of complete automation, Otsu's method adaptively calculates the optimum threshold by maximizing the inter-class variance. However, an interpreter can also interactively fine tune the adaptive threshold to highlight the regions around salt body, which results in better delineation of salt domes. The output of thresholding is shown in Figure~\ref{SS160_Binmap}, which highlights the regions around salt-dome boundary i.e. the most salient area in the saliency map.

\subsection{Region Growing}
Thresholding yields a volume $\mathbf{B}$, same size as that of the 3D seismic volume, $\mathbf{V}$, which contains noisy and disconnected regions as evident in a seismic inline shown in Figure~\ref{SS160_Binmap}. In order to extract a salt body from binary volume $\mathbf{B}$, we apply 3D region growing method as a third step to obtain a closed salt body. In region growing, we randomly select either one or multiple seed points inside salt dome and keep adding to each seed point a set of neighboring voxels until they hit the salt boundary. \citealt*{RG} proposed a robust, rapid, and free-of-parameter-tuning region growing method for the segmentation of intensity images. In region growing, we select multiple seed points $p_{s1}$, $p_{s2}$, $p_{s3}$, ... $p_{sr}$ and keep adding to each seed point a set of neighboring voxels until a stopping criterion is met. Voxels that meet a certain criterion form regions $R_1$, $R_2$, $R_3$, ... $R_r$ and are labeled as allocated voxels. In contrast, voxels in the variant regions that do not meet selected criteria are labeled as unallocated. We define $U$ as the set of all unallocated voxels in the neighborhood of labeled regions or salt body, which can be mathematically expressed as

\begin{figure}[t]
  \centering
  \includegraphics[width=.7\columnwidth]{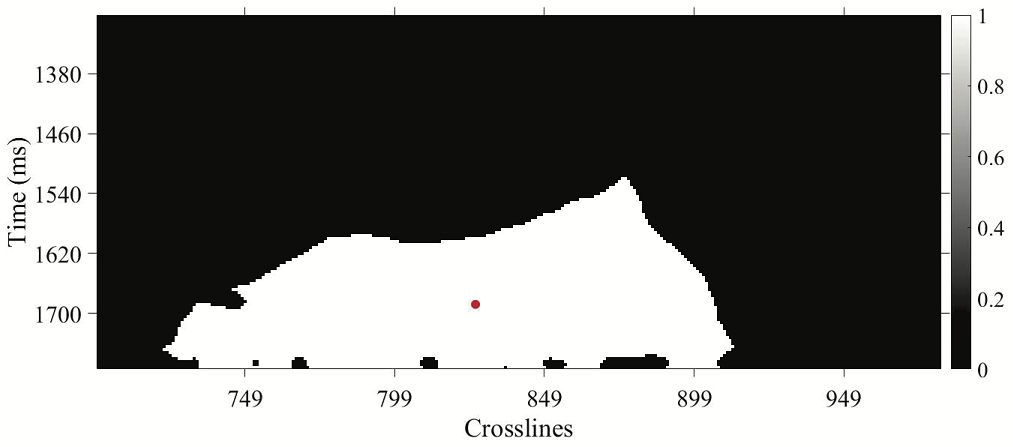}
  \caption{The output of 3D region growing with a seed point highlighted in red.}\label{SS160_RG}
\end{figure}

\begin{equation}
\label{equ:rg}
% U = \left\{ v \notin \bigcup_{i=1}^{r} R_i|N(v)\cap \bigcup_{i=1}^{r} R_i\neq\phi \right\},
U = \left\{ v \suchthat v \notin \bigcup_{i=1}^{r} R_i \suchthat N(v)\cap \bigcup_{i=1}^{r} R_i  \neq\phi \right\},
\end{equation}
where a voxel at point $[m,n,k]$ is represented as $v$ for simplification, $\phi$ represents the empty set, and $N(v)$ is the neighborhood of voxel $v$ in the 3D volume. In equation (\ref{equ:rg}), we compute the intersection of neighboring voxels, $N(v)$, and the union of all allocated regions, $R_i$. If this intersection is not an empty set, then $U$ contains all the voxels that do not lie inside allocated regions union. For unlabeled voxels $v \in U$, the $N(v)$ falls within just one of the labeled regions $R_i$. We define $\psi(v)\in \{1,2,...,r\}$ as the indexes of neighboring voxels such that $N(v) \cap \psi(v) \neq \phi$. In intensity-based region growing, voxels are assigned to particular regions, $R_i$, based on their intensity values, $I(v)$. $\delta(v)$ define the intensity difference at voxel $v$ and its adjacent labeled region $R_i$ as
\begin{equation}
\label{equ:delta}
\delta(v) = | I(v)- mean_{n \in R_i}(I(n))|,
\end{equation}
where $I(v)$ defines the intensity values at the voxels $v$. If $N(v)$ is close to more than one $R_i$, then $\psi(v)$ takes the value of $v$ such that $\delta(v)$ at $R_i$ is minimized.
\begin{equation}
\label{equ:minimization}
\psi(v) = \min_{v\in U}\{ \delta(v)\}.
\end{equation}

In region growing, we select an initial seed point, $p_s$, randomly inside the salt body and it continues to grow into a region until it hits the highlighted boundary (labeled as \emph{ones} in $\mathbf{B}$). The seeded region growing continues until all the voxels are allocated to region $R_i$. This process can be mathematically expressed as follows:
\begin{equation}
\label{equ:BRG}
SD = \left\{ \underset{p_g\in\mathbf{B}}{\bigcup}R|N(p_{g}) \neq 1 \right\},
\end{equation}
where $N(p_g)$ is the neighboring region of the area starting from the seed point $p_s$ and $SD$ is the salt-dome body detected after region growing. In the nutshell, we apply region growing on the binarized volume, $\mathbf{B}$, obtained by thresholding saliency map, $\mathbf{S}$, to yield a salt-dome body, $SD$, with a seed point selected randomly inside salt body. If we want to segment multiple disconnected salt bodies then we can select multiple seed points that will independently grow into salt dome. Our dataset (details are given in experimental results section) has only one salt dome and we have randomly selected only one seed point for whole volume to initiate 3D region growing. The output of 3D region growing with an initial seed point highlighted in red is shown in Figure~\ref{SS160_RG}.

\subsubsection{Seed Point Selection}
The seed point, $p_s$, for region growing can be selected either automatically by computer algorithms or manually by the seismic interpreter. The automatic seed point selection methods for 2D and 3D data, based on directionality and tensor decomposition, are given in \citealt*{Zhen2015GoT} and \citealt*{Shafiq2016_GoT_Intr}, respectively. However, automatic seed point selection methods are computationally expensive and may fail in the presence of noise and chaotic horizons. Automatic seed point selection methods under such circumstances may result in the considerable loss of time and computation effort. In manual seed point selection, the seismic interpreter can interactively choose \emph{any} arbitrarily random point inside volume as long as it is inside salt body. The interpreter can also choose multiple seed points to speed up the region growing. The time required by geophysicist interpreter to manually select $p_s$ is insignificant as compared to automatic $p_s$ selection. Therefore, in this paper, we have manually selected a seed point inside salt body.

\begin{figure}[t]
  \centering
%  \vspace{0.7cm}
  \includegraphics[width=.7\columnwidth]{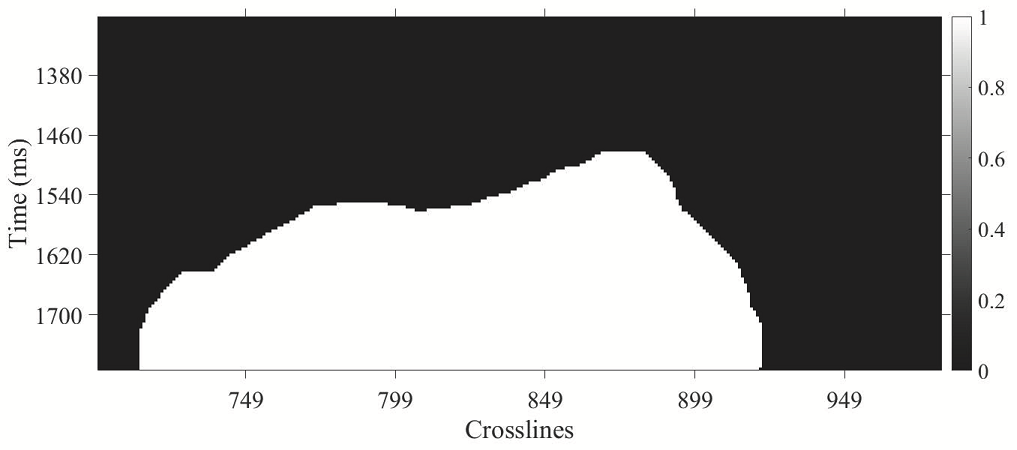}
  \caption{Dilated salt dome.}\label{SS160_PP}
\end{figure}

\subsection{Post-Processing}

\begin{figure}
  \centering
  \includegraphics[width=.7\columnwidth]{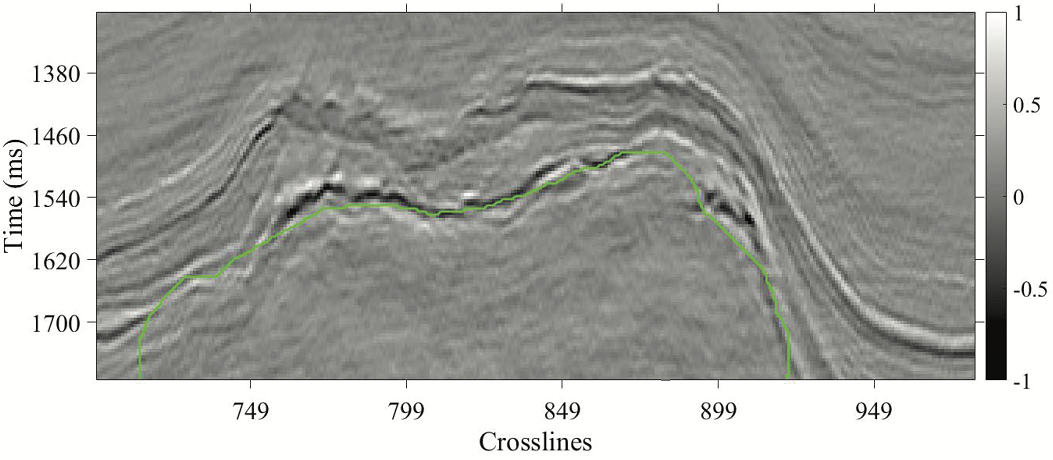}
  \caption{The green line depicts the output of post-processing.}\label{SS160_delin}
\end{figure}

To bridge the gaps between the output of region growing and the salt-dome boundary, we apply morphological operations, which includes dilation and perimeter extraction. By expanding the detected salt body, the dilation operation matches the detected salt-dome boundary with the reference as closely as possible and alleviate the effects of window sizes in the calculation of the saliency. The dilated salt body is mathematically given as
\begin{equation}
\label{equ:sd}
SD_D =  SD \oplus H_D,
\end{equation}
where $\oplus$ represents dilation and $H_D$ represents the structural element of dilation (Gonzalez and Woods 2008). The final step of the proposed workflow is to detect boundary by extracting the perimeter of the salt body, which can be mathematically expressed as
\begin{equation}
\label{equ:sdb}
SD_B =  \left \{ v \in SD_D|N_{26}[v] - SD_D \neq \phi \right \},
\end{equation}
where $SD_B$ is the detected salt-dome boundary and $N_{26}[v]$ defines the 26 neighboring voxels of $v$ in a 3D space. The dilated salt-dome and the output of post-processing are shown in Figures~\ref{SS160_PP} and \ref{SS160_delin}, respectively.

\section*{Experimental Results}
\label{sec:results}
\begin{figure}[t]
\centering
  \begin{minipage}[b]{.7\linewidth}
  \centering
  \centerline{\includegraphics[width=\columnwidth]{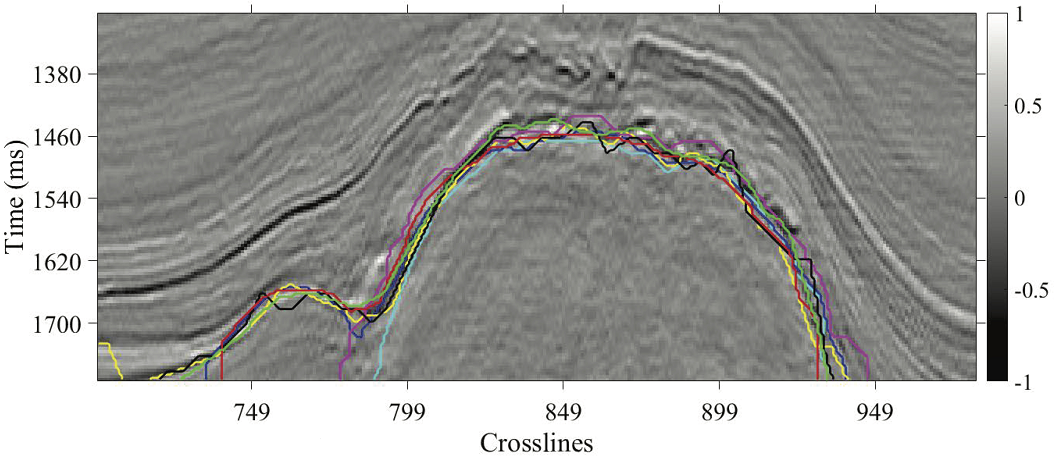}}
  \centerline{(a)}
\end{minipage}
\\
\begin{minipage}[b]{.7\linewidth}
  \centering
  \centerline{\includegraphics[width=\columnwidth]{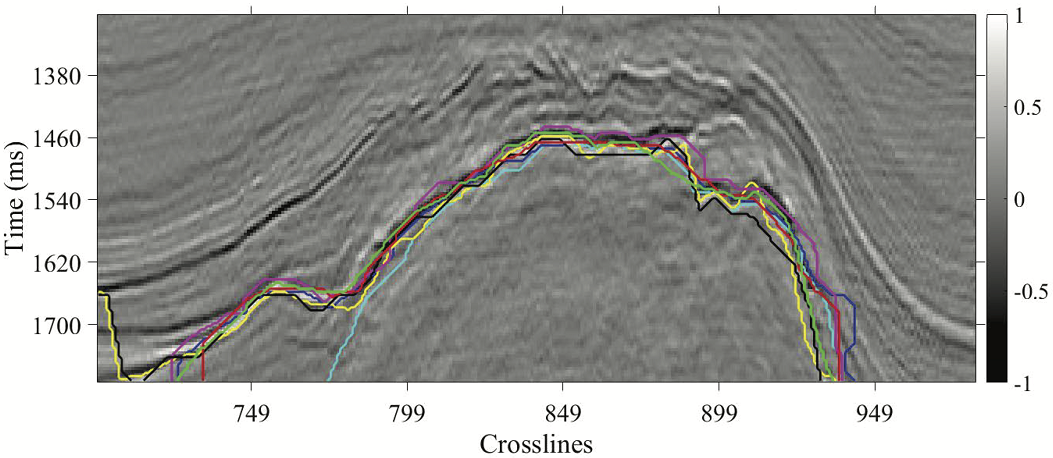}}
  \centerline{(b)}
\end{minipage}
\\
\begin{minipage}[b]{.7\linewidth}
  \centering
  \centerline{\includegraphics[width=\columnwidth]{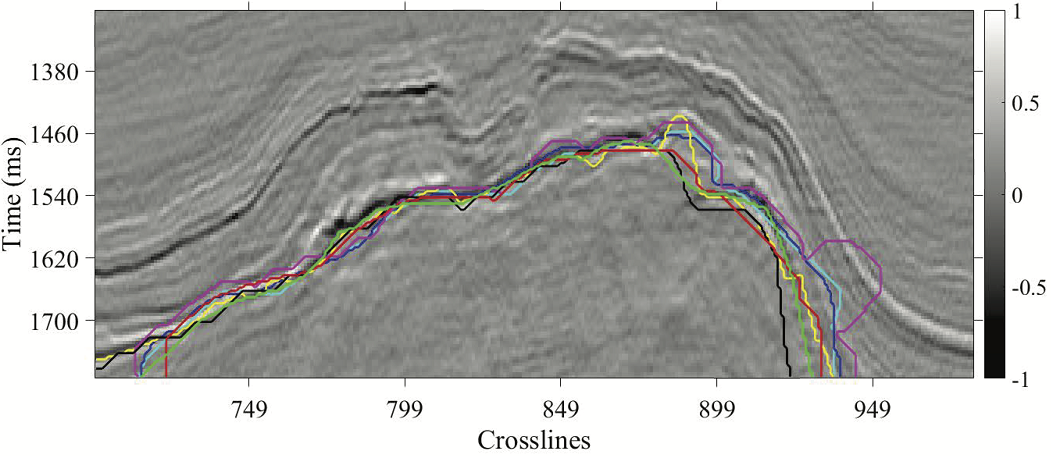}}
  \centerline{(c)}
\end{minipage}
\\
\begin{minipage}[b]{.7\linewidth}
  \centering
  \centerline{\includegraphics[width=\columnwidth]{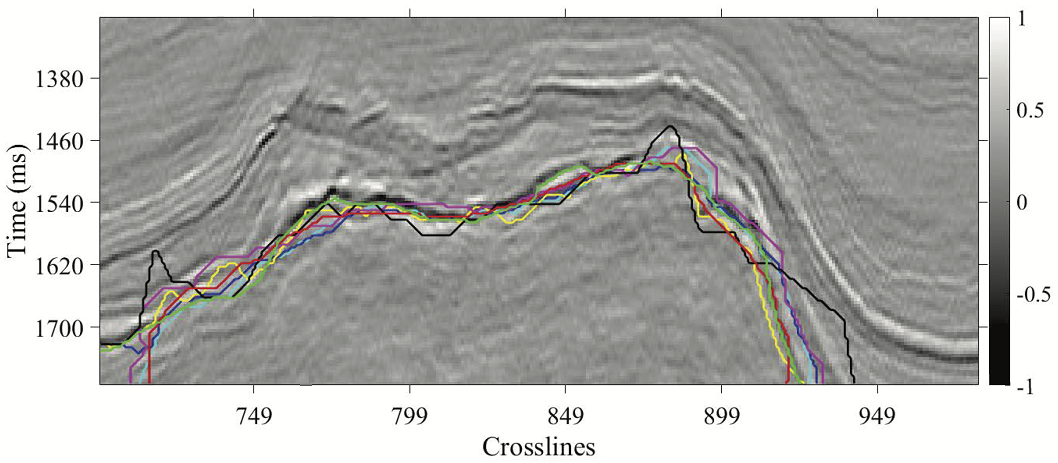}}
  \centerline{(d)}
\end{minipage}
\caption{The experimental results of salt dome delineation on different seismic inline sections. Green: Reference boundary manually interpreted by a geophysicist, Cyan:~\citealt*{Zhen2015GoT}, Blue:~\citealt*{Shafiq2015GoT}, Yellow:~\citealt{berthelot2013texture}, Magenta:~\citealt{aqrawi2011detecting}, Black:~\citealt*{asjadcodebook}, Red: Proposed Workflow. (a) Seismic inline section 360, (b) Seismic inline section 372, (c) Seismic inline section 390, (d) Seismic inline section 408.}
\label{Sal_op}
\end{figure}

In this section, we demonstrate the effectiveness of the proposed workflow for salt dome delineation. We have used the real seismic dataset acquired from the Netherlands offshore, $F3$ block in the North Sea whose size is $24$ x $16$ $km^2$ (dGB Earth Sciences B.V. 1987). The seismic volume that contains the salt-dome structure has an inline number ranging from $151$ to $501$, a crossline number ranging from $701$ to $981$, and a time direction starting from $1,300ms$ sampled every $4ms$. The bin size across the inline and crossline directions is 25 meters. In this paper, we use a non-overlapping cube of size $3 \times 3 \times 3$ for saliency calculation. The size of structuring element in the post-processing operations is equal to the side length of the cube i.e. $3$. The output of the proposed workflow and the results of different algorithms for salt dome delineation on seismic inline sections 360, 372, 390, and 408 are shown in Figure~\ref{Sal_op}, with the reference boundary manually labeled by a geophysicist in green. The cyan, blue, yellow, magenta, black, and red lines represent the boundaries detected by \citealt*{Zhen2015GoT}, \citealt*{Shafiq2015GoT}, \citealt{berthelot2013texture}, \citealt{aqrawi2011detecting}, \citealt*{asjadcodebook}, and the proposed workflow, respectively.

The output of different state-of-the-art algorithms have significant differences from the reference boundary as observed in the Figure~\ref{Sal_op}. The output of the texture-based method by \citealt{berthelot2013texture} and learning-based method by \citealt*{asjadcodebook} extends beyond the salt-dome boundary as seen in the bottom left corner of Figure~\ref{Sal_op}a-b. The edge-based method by \citealt{aqrawi2011detecting} deviates from the reference in the absence of strong seismic reflections as observed in the bottom right section of salt dome in Figure~\ref{Sal_op}c. Furthermore, edge-based method by \citealt{aqrawi2011detecting} and texture-based method by \citealt*{Zhen2015GoT} detects only the right side of salt dome and is not able to detect the associated event on the left side as observed in Figure~\ref{Sal_op}a-b. The output of texture-based methods presented in \citealt*{Zhen2015GoT} and \citealt*{Shafiq2015GoT} degrades in the absence of strong texture as seen in Figure~\ref{Sal_op}c. The proposed method captures the salient areas in the volume and attempt to highlight the variations of salt dome across seismic volume. However, this method yields less accurate results in the areas where the variations in salt-dome boundary are comparatively less as compared to their neighboring seismic inlines. As shown in Figure~\ref{Sal_op}a-c, the proposed method diverges from the reference boundary in the bottom left areas of salt dome. Subjectively, it can be argued that the boundaries detected by all aforementioned algorithms lie close to the reference boundary. However, due to tortuous salt-dome boundaries, it is difficult to deduce which method outperforms other methods in terms of its delineation precision and accuracy. Therefore, we have used five different metrics to objectively evaluate the results of different salt-dome delineation algorithms.

\begin{figure}[t]
\centering
  \begin{minipage}[b]{.7\linewidth}
  \centering
  \centerline{\includegraphics[width=\columnwidth]{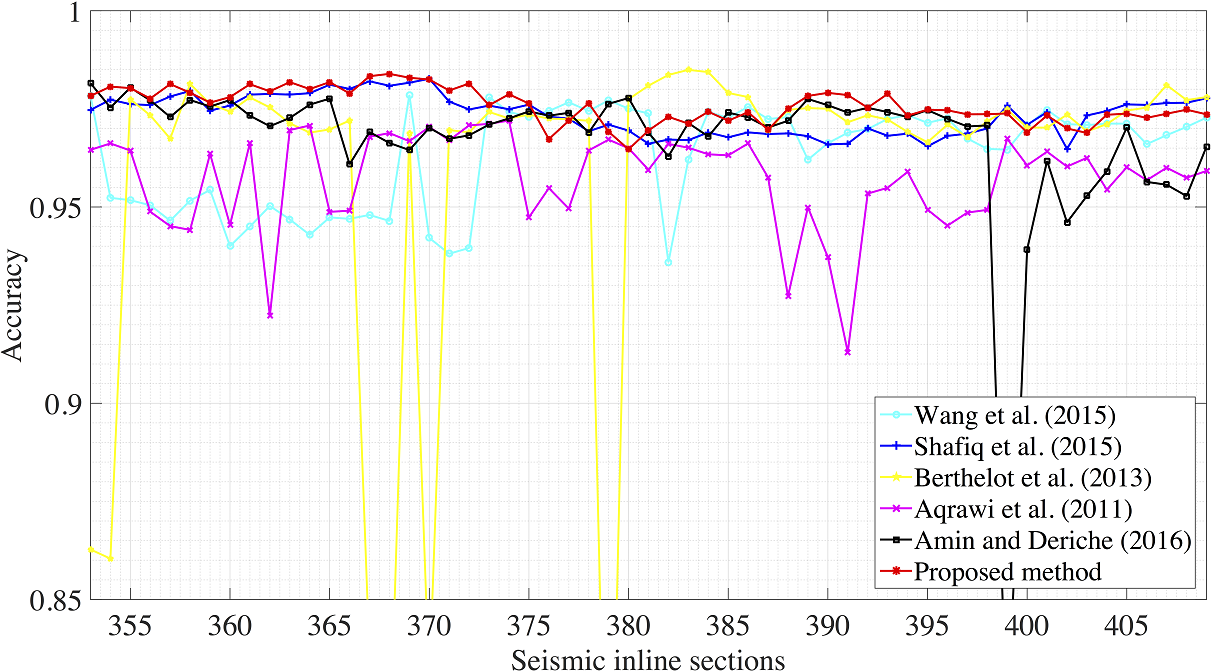}}
  \centerline{(a)}
\end{minipage}
\\
\begin{minipage}[b]{.7\linewidth}
  \centering
  \centerline{\includegraphics[width=\columnwidth]{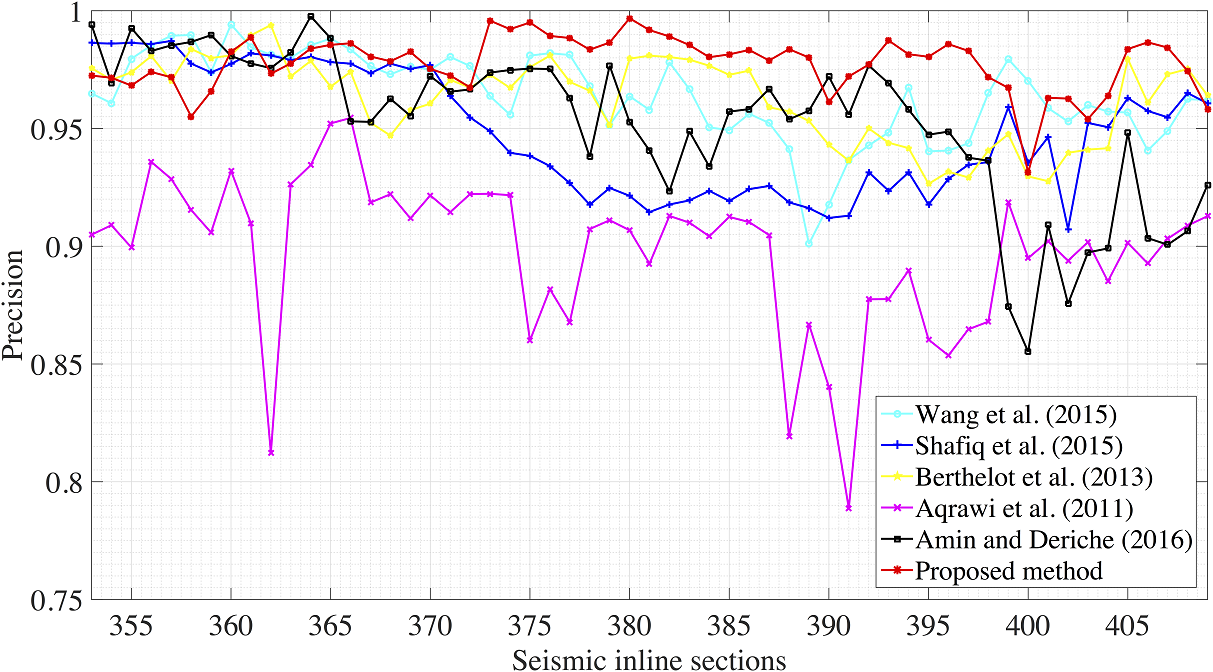}}
  \centerline{(b)}
\end{minipage}
\\
\begin{minipage}[b]{.7\linewidth}
  \centering
  \centerline{\includegraphics[width=\columnwidth]{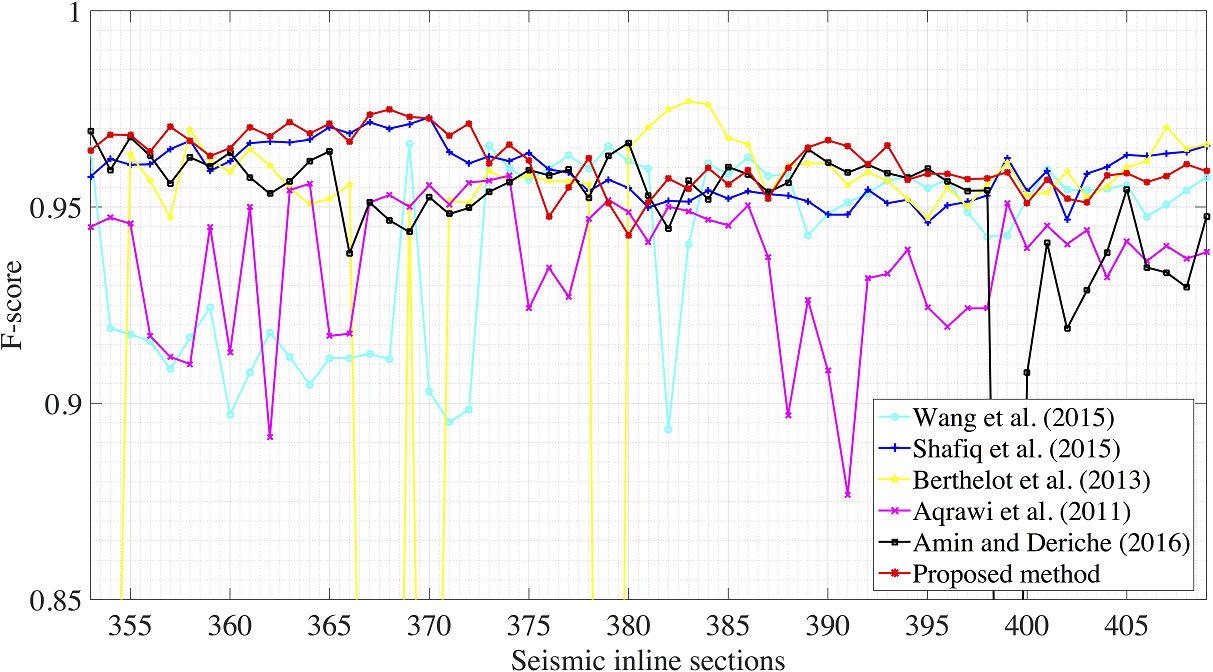}}
  \centerline{(c)}
\end{minipage}
\caption{Accuracy, precision, and F-scores of various salt-dome delineation algorithms for fifty seven consecutive seismic inline sections. (a) Accuracy, (b) Precision, (c) F-score.}
\label{ObjectiveMeasure}
\end{figure}

To investigate the results of delineation algorithms per pixel, we have used accuracy, precision, and F-score (Powers 2011), which are usually used as an objective evaluation metrics in binary classification. True positives (TP) and true negatives (TN) measures, at each seismic inline, the number of pixels that belong to the salt-dome and are correctly identified as such and vice versa. On the other hand, false positives (FP) and false negatives (FN) measures, at each seismic inline, the number of non-salt pixels classified as salt pixels and salt pixels classified as non-salt pixels, respectively. Accuracy, precision, and F-score are then calculated using
\begin{align}\label{eqn:APF}
& Accuracy = \frac{TP + TN}{TP + FN + FP + TN}, \\
& Precision = \frac{TP}{TP + FP}, \\
& Recall = \frac{TP}{TP + FN},\\
& \Fscore = 2 \cdot \frac{Precision \cdot Recall}{Precision + Recall}.
\end{align}
The accuracy and precision of various delineation algorithms for fifty seven consecutive seismic inlines are shown in Figures~\ref{ObjectiveMeasure}a and \ref{ObjectiveMeasure}b, respectively. The accuracy and precision values of the proposed method are closer to one that not only support that it is more accurate but also more precise in delineating salt-domes within seismic inlines. F-score, which measures the test accuracy, is the harmonic mean of precision and recall. F-scores of various delineation algorithms for fifty seven consecutive seismic inlines are shown in Figure~\ref{ObjectiveMeasure}c, which demonstrates that the proposed workflow, on most inlines, surpass other methods of salt dome delineation. The mean and standard deviation (S.D) of accuracy, precision, and F-scores of various salt-dome delineation algorithms for the plots shown in Figure~\ref{ObjectiveMeasure} are presented in Table~\ref{Table_ObjSimClass}, with the best scores highlighted in boldface. It can be observed that the proposed workflow outperforms other salt-dome delineation algorithms in terms of mean and standard deviation, except the standard deviation of F-score, which is slightly less than the texture-based method by \citealt*{Shafiq2015GoT}.

\begin{table}[t]
\caption{Mean and standard deviation of accuracy, precision, and F-scores for various salt-dome delineation algorithms.}\label{Table_ObjSimClass}
\begin{center}
\begin{tabular}{|l||l|l||l|l||l|l||}
\hline
\multirow{2}{*}{~~~~~~~~\textbf{Methods}}            &\multicolumn{2}{l||}{Accuracy (\%)}      &   \multicolumn{2}{l||}{Precision (\%)}   &   \multicolumn{2}{l||}{~~~~~F-score} \\
\cline{2-7}
                                               & Mean     & S.D.        & Mean    & S.D.        & Mean   & S.D.                 \\
\hline\hline                                    % Accuracy              % Precision             % F-measure
\citealt*{Zhen2015GoT}                  & 96.32  & 1.30         & 96.40  & 1.85         & 0.9403 & 0.0232                \\
\citealt*{Shafiq2015GoT}              & 97.35  & 0.50         & 94.86  & 2.59         & 0.9591 & \textbf{0.0070}       \\
\citealt{berthelot2013texture}    & 95.84  & 4.75         & 96.26  & 1.77         & 0.9239 & 0.1085                \\
\citealt{aqrawi2011detecting}        & 95.72  & 1.22         & 89.79  & 3.20         & 0.9361 & 0.0176                 \\
\citealt*{asjadcodebook}           & 96.69  & 2.23         & 95.26  & 3.24         & 0.9470 & 0.0447                 \\
Proposed Workflow                               & \textbf{97.59}
                                                & \textbf{0.45}         & \textbf{97.76}
                                                                        & \textbf{1.19}         & \textbf{0.9616} & 0.0072      \\
\hline
\end{tabular}
\end{center}
\end{table}

\begin{figure}[t]
\centering
  \begin{minipage}[b]{.715\linewidth}
  \centering
  \centerline{\includegraphics[width=\columnwidth]{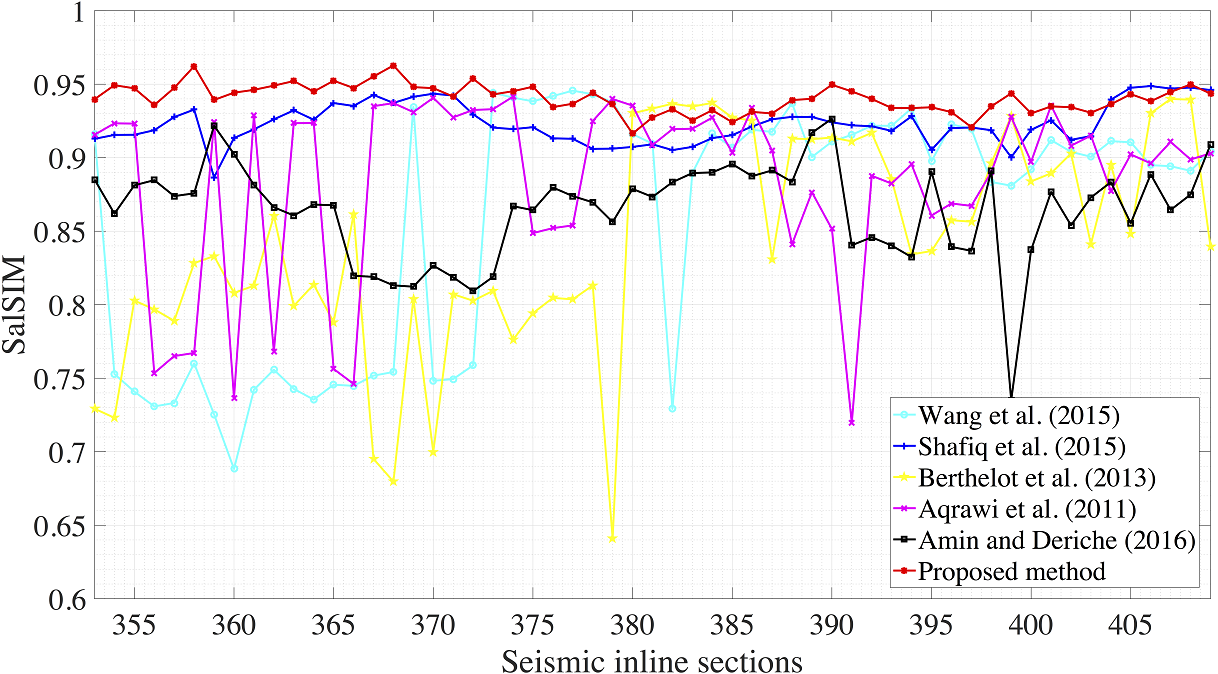}}
  \centerline{(a)}
\end{minipage}
\\
\begin{minipage}[b]{.7\linewidth}
  \centering
  \centerline{\includegraphics[width=\columnwidth]{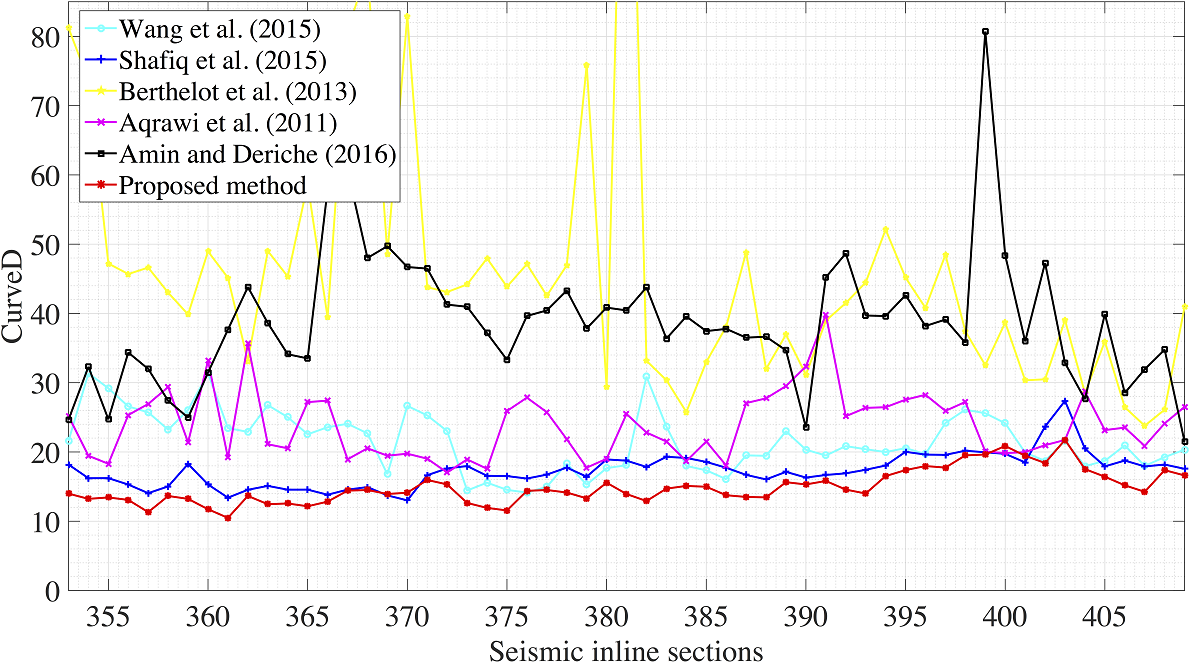}}
  \centerline{(b)}
\end{minipage}
\caption{\emph{SalSIM} and \emph{CurveD} indices of various salt-dome delineation algorithms for fifty seven consecutive seismic inline sections. (a) \emph{SalSIM}, (b) \emph{CurveD}.}
\label{Obj_Eval_SC}
\end{figure}

Accuracy, precision, and F-scores provided the statistical evaluation of results in terms of pixels. To evaluate the results of different salt-dome delineation algorithms based on their shape and curvature, we have used two different evaluation metrics. First, \emph{SalSIM}, proposed by \citealt*{Zhen2015GoT}, measures the similarity between the reference and the detected salt-dome boundary using the Fr{\'e}chet distance-based similarity index. The \emph{SalSIM} index ranges between 0 to 1, with a higher value indicating a greater similarity between the two boundaries under comparison. Second, \emph{CurveD}, proposed by \citealt*{Shafiq2016_GoT_Intr}, computes the distance between the reference and the detected salt-dome boundary based on their shape and curvedness. If two curves are similar in shape and curvature then the \emph{CurveD} index is closer to zero and vice versa. The \emph{SalSIM} and \emph{CurveD} indices of various salt-dome delineation algorithms for fifty seven consecutive seismic inlines are shown in Figures~\ref{Obj_Eval_SC}a and \ref{Obj_Eval_SC}b, respectively. Furthermore, the mean and standard deviation (S.D.) of \emph{SalSIM} and \emph{CurveD} indices, illustrated in Figure~\ref{Obj_Eval_SC}, are given in Table~\ref{Table_ObjSim_SC}, with the best scores highlighted in boldface. Figure~\ref{Obj_Eval_SC} and Table~\ref{Table_ObjSim_SC} illustrate that the best results are obtained using the proposed method, which outperforms the state-of-the-art methods for salt dome delineation.

\begin{figure}[t]
  \centering
  \includegraphics[width=.7\columnwidth]{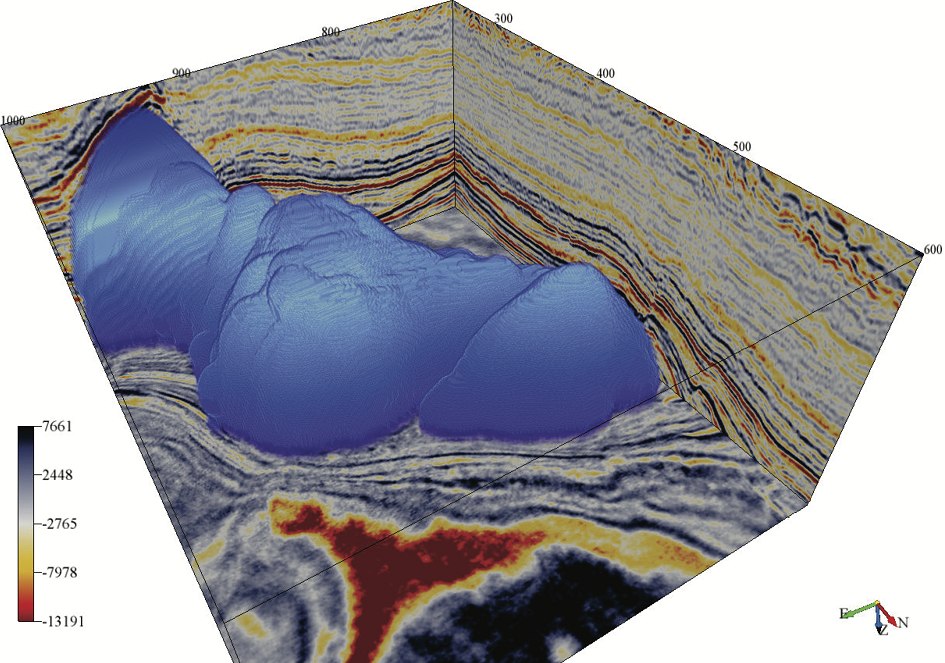} %
  \caption{The detected 3D salt body.}\label{3D_SaltBody}
\end{figure}

We also calculate the time required by each algorithm for delineation and results are summarized in Table~\ref{Table_ObjSim_SC}. It can be observed that the proposed workflow is not only computationally very efficient but also the fastest among all aforementioned algorithms. Finally, the detected 3D salt body is displayed in Figure~\ref{3D_SaltBody}, which illustrates that the proposed method not only outlines the major structure of the salt body but also highlight the details of local structures effectively to show the formation of salt dome. Experimental results presented in this section show that the proposed workflow can be a useful addition to the interpreters toolbox for delineating important geological structures.

\begin{table}[t]
\caption{Mean and standard deviation of \emph{SalSIM} and \emph{CurveD} indices, and time required by various salt-dome delineation algorithms.}\label{Table_ObjSim_SC}
\begin{center}
\begin{tabular}{|l||l|l||l|l||l|}
\hline

\multirow{2}{*}{~~~~~~~~\textbf{Methods}}       &   \multicolumn{2}{l||}{~~~~~~\emph{SalSIM}} &   \multicolumn{2}{l||}{~~~~~~~\emph{CurveD}}    &  \multirow{2}{*}{Time(s)}          \\
\cline{2-5}
                                                & Mean    & S.D.                        & Mean              & S.D.                              &                           \\
\hline\hline                                    % SalSIm                                % CurveD
\citealt*{Zhen2015GoT}                  & 0.8573  & 0.0844                      & 21.5004           & 4.2859                            &  11.4895                  \\
\citealt*{Shafiq2015GoT}              & 0.9232  & 0.0136                      & 17.3355           & 2.4872                            &  63.3162                  \\
\citealt{berthelot2013texture}    & 0.8439  & 0.0730                      & 45.9306           & 19.7698                           &  33.5447                  \\
\citealt{aqrawi2011detecting}        & 0.8845  & 0.0605                      & 23.8682           & 4.8281                            &  0.98110                  \\
\citealt*{asjadcodebook}           & 0.8643  & 0.0333                      & 38.7502           & 9.6533                            &  0.68480                  \\
Proposed Workflow                               & \textbf{0.9405}& \textbf{0.0095}      & \textbf{14.8835}  & \textbf{2.4268}                   &  \textbf{0.39520}          \\
\hline
\end{tabular}
\end{center}
\end{table}

\section*{Conclusion}
\label{sec:conclusion}
A seismic attribute based on visual saliency, \emph{SalSi}, has many applications in seismic interpretation such as salt dome delineation, tracking salt domes in a seismic volume, algorithms initialization, reducing time computation, seismic retrieval, and labeling, etc. In this paper, we proposed a workflow based on \emph{SalSi} for salt dome delineation within migrated seismic volumes. The experimental results on a real seismic dataset from the North Sea, F3 block show the effectiveness of the proposed workflow. The subjective and objective evaluation of the results show that the proposed workflow is very fast and outperforms the state-of-the-art algorithms for salt dome delineation. The results presented in this paper show a promising future of the proposed workflow for salt dome delineation and excellent potential in seismic interpretation that can not only automate but can effectively reduce the time as well. This workflow can also be easily modified to highlight chaotic horizons and faults within seismic volumes, which requires further investigation.

\section{Acknowledgments}
This work is supported by the Center for Energy and Geo Processing (CeGP) at the Georgia Tech and KFUPM. We would like to thank dGB Earth Sciences for making the F3 seismic data available publicly.

\newpage

%\bibliographystyle{gp}  % style file is seg.bst
% \bibliography{refs}

%\newpage

%\listoffigures

%\newpage

%\listoftables

\end{document}